\documentclass[runningheads]{llncs}
\usepackage[T1]{fontenc}
\usepackage{graphicx}
\usepackage{multirow}
\usepackage{graphicx,verbatim}
\usepackage{amsmath}
\usepackage{amssymb}
\usepackage{hyperref}
\usepackage{siunitx}
\usepackage{tikz}
\usetikzlibrary{backgrounds}
\usepackage{booktabs}
\usepackage{placeins}
\usepackage[normalem]{ulem}
\newcommand{\bftab}{\fontseries{b}\selectfont}

\usepackage{color}

\begin{document}
\title{Field-of-View Extension in Dental Cone-Beam CT via Implicit Neural Representations and Diffusion Model-Based Refinement}
\titlerunning{FOV Extension in Dental CBCT}
%
\author{Susanne Schaub\inst{1}\textsuperscript{*}\orcidID{0009-0008-6278-4855}\and
Florentin Bieder\inst{1}\orcidID{0000-0001-9558-0623} \and
Matheus L. Oliveira\inst{2}\orcidID{0000-0002-8054-8759} \and
Yulan Wang\inst{3}\orcidID{0000-0003-0502-6931} \and
Buyanbileg Sodnom-ish\inst{4}\orcidID{0000-0002-4239-1420} \and
Dorothea Dagassan-Berndt\inst{5}\orcidID{0000-0003-1109-8763} \and
Michael M. Bornstein\inst{4}\orcidID{0000-0002-7773-8957} \and
Philippe C. Cattin\inst{1}\orcidID{0000-0001-8785-2713}
}
\authorrunning{S. Schaub et al.}
%
\institute{Department of Biomedical Engineering, University of Basel, Allschwil, Switzerland \\
\email{\textsuperscript{*}Corresponding author(s). E-mail(s):  s.schaub@unibas.ch}\\ \and
Piracicaba Dental School, University of Campinas, Piracicaba, Brazil \and
State Key Laboratory of Oral \& Maxillofacial Reconstruction and Regeneration,
Key Laboratory of Oral Biomedicine Ministry of Education, Hubei Key Laboratory of
Stomatology, School \& Hospital of Stomatology, Wuhan University, Wuhan, China \and
Department of Oral Health \& Medicine, University Center for Dental Medicine,
University of Basel, Basel, Switzerland \and
Dental Imaging, University Center for Dental Medicine, University of Basel, Basel, Switzerland \\
}

\maketitle              
\begin{abstract}
Dental cone-beam computed tomography (CBCT) systems often employ detector
configurations that provide a truncated field of view (FOV) that only captures a small part of the patient's anatomy. In this work, we aim to reconstruct an extended FOV using projections of truncated FOV scans. To this end, we propose a three-stage framework that consists of (1) an implicit neural representation (INR) for estimating missing parts of the truncated projection data, (2) an iterative reconstruction for generating a secondary volumetric image with improved anatomical consistency and (3) a fast diffusion model for image enhancement. The proposed approach combines the strengths of continuous representations, physics-based reconstruction and generative refinement within a unified pipeline for truncated CBCT imaging. Experimental results demonstrate that the method effectively reduces truncation artifacts, improves the reconstruction of structures extending beyond the original FOV and produces images with enhanced quality. Our code is
publicly available at \url{https://github.com/SusanneSchaub/CBCT-FOV-Extension}.

\end{abstract}

\section{Introduction}
In dental cone-beam computed tomography (CBCT), compact and low-cost scanner configurations commonly rely on small detectors, yielding a limited field of view (FOV) that may not include the entire anatomy of the patient’s head. Furthermore, the use of a small FOV in CBCT is recommended whenever clinically feasible, as it indirectly helps to reduce the radiation dose received by the patient. However, restricting the FOV results in missing projection data, rendering the reconstruction problem ill-posed. 

A wide range of approaches have been developed to address the problem of FOV extension using truncated projection data. Classical methods do sinogram completion analytically or by fitting simple shapes to enforce projection consistency and then use the FDK-algorithm~\cite{feldkamp1984practical} or iterative reconstruction techniques to reconstruct the volumetric images~\cite{hsieh2004novel}. Iterative reconstruction methods with several different regularizations have been proposed for CT reconstruction~\cite{tian2011low}. However, applying iterative reconstruction approaches to truncated projections leads to an incomplete forward model, leaving attenuation values outside the measured FOV weakly constrained. As a result, reconstruction errors associated with the missing data propagate throughout the iterative optimization, producing truncation artifacts and degrading image quality.

Alongside these classical methods, multiple learning-based techniques have emerged. Since CBCT generates a large amount of projection data, methods operating in the projection domain are often computationally expensive and memory-intensive. To mitigate this challenge, image-to-image-translation methods are typically adopted on a slice-wise basis~\cite{huang2021data,liman2024diffusion,xu2023body}. Moreover, learning-based approaches that operate in both the projection and image reconstruction domains have also been proposed~\cite{gao2023transformer,han2026cone}.
\begin{figure}[t]
    \centering
    \includegraphics[width=\textwidth]{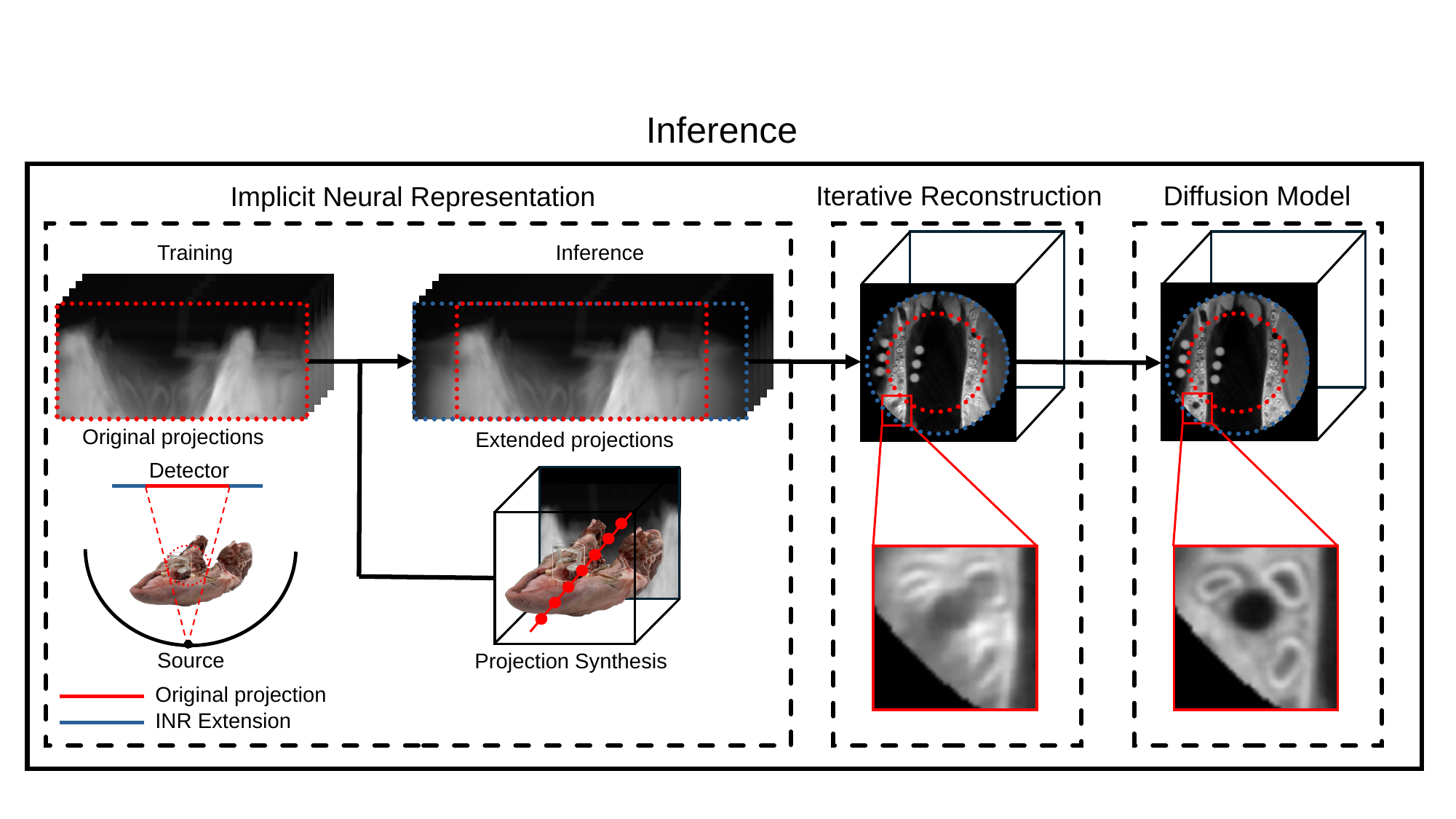}
    \caption{An INR first extends the FOV of truncated projections, followed by iterative reconstruction and diffusion-based refinement to recover lost detail.}
    \label{fig:inference}
\end{figure}

Diffusion models have proven to be capable of generating state-of-the-art results in generative modeling and outperform generative adversarial networks. These models have been applied to address several challenges in CBCT imaging, including limited-angle CBCT reconstruction~\cite{xie2024prior}. Some recent works~\cite{li2026efficientimagetoimageschrodingerbridge,liman2024diffusion} leverage these models to recover the CT FOV.

More recently, implicit neural representations (INRs) have emerged as a promising approach for addressing ill-posed inverse problems, including sparse-view CBCT reconstruction~\cite{zang2021intratomo,zha2022naf}. INRs are used for CBCT reconstruction by formulating the problem as the optimization of a continuous attenuation field. This continuous representation allows the attenuation field to be evaluated at arbitrary spatial locations, removing the constraint of a fixed voxel resolution. Since the entire attenuation field is parameterized by a single neural network, each optimization step updates a globally consistent representation influenced by all projection views. These properties make INRs well suited for CBCT FOV extension, as they reduce truncation artifacts and enable estimation of attenuation values outside the measured FOV while maintaining consistency with the acquired projection data. However, the application of INRs to truncated-FOV extension remains largely unexplored. Existing work has shown that incorporating INR-based coarse-grid
priors into an iterative reconstruction framework can mitigate truncation artifacts and improve the visibility of low-contrast structures compared with conventional iterative reconstruction methods that do not employ a coarse-grid prior~\cite{park2025iterative}. 

In this work, we propose a three-stage approach consisting of INR-based projection completion, iterative image reconstruction and diffusion model-based image refinement. The inference process is shown in Fig.~\ref{fig:inference}.

\section{Methods}
\subsection{Implicit Neural Representations for CBCT Reconstruction}
INRs are used for CBCT reconstruction by modeling the attenuation field as a continuous function parameterized by a neural network that maps spatial coordinates to attenuation coefficients $\mu$. The spatial coordinates are positionally encoded, and the network is trained using projection images acquired from multiple viewing angles. The X-ray intensity is described by Beer’s law:
\begin{equation}
I = I_0 \exp\left(-\sum_{i=1}^{N} \mu_i \delta_i\right)
\end{equation}
where $I_0$ is the initial intensity and $\delta_i$ the distance between adjacent samples along the ray $\mathbf{r}$. Given the CBCT imaging geometry, the predicted attenuation field is integrated along each ray to obtain the synthetic projection $I_s$. The network parameters are optimized by minimizing the discrepancy between the synthetic projections $I_s$ and the measured projections $I_r$ for a batch of rays $B$:
\begin{equation}
\mathcal{L} = \sum_{\mathbf{r} \in B} \left\| I_r(\mathbf{r}) - I_s(\mathbf{r}) \right\|^2.
\end{equation}
\subsection{Pipeline}
\label{sec:pipeline}
First an INR was trained to overfit the truncated projection data following~\cite{zha2022naf}. We extended this implementation with three modifications: (i) a total variation (TV) regularization term to encourage spatial smoothness and suppress high-frequency fluctuations. The network predicts a randomly shifted $32^3$ voxel grid $V$, on which spatial gradients along the three coordinate axes are penalized:
\begin{equation}
\mathcal{L}_{\text{TV}}=\lambda \sum_{(i,j,k) \in \Omega} \sqrt{ (V_{i+1,j,k}-V_{i,j,k})^2 + (V_{i,j+1,k}-V_{i,j,k})^2 + (V_{i,j,k+1}-V_{i,j,k})^2},
\end{equation}
where we set $\lambda=10^{-5}$ empirically.
Further, (ii) we add a normalization layer between the hash-grid encoding and the neural network~\cite{xu2025nerf} and (iii) a hash-grid encoding regularization to improve reconstruction quality following~\cite{shin2023fast}. After training, the INR was used to render extended projections corresponding to an enlarged FOV. These projections were subsequently reconstructed into a volumetric image using the iterative \emph{OS-SART} algorithm~\cite{censor2002block}. In both datasets, we reconstructed volumes of shape $256 \times 256 \times 256$. For the MMDental dataset, the voxel size was set to \SI{0.9}{\milli\metre}, while for the pig jaw dataset it was \SI{1.0}{\milli\metre}.

Finally, a conditional diffusion model was applied to recover fine anatomical details. The architecture of the diffusion model is based on the 2D denoising diffusion probabilistic model (DDPM) approach proposed in~\cite{nichol2021improved}. To accelerate inference, we introduced several modifications, including a variance-preserving noise schedule with a linearly varying noise intensity and the use of an $L_1$ loss, following the approach of~\cite{durrer2025fastwdm3d}.
We set the number of diffusion steps to three based on empirical evaluation of different numbers of diffusion steps. We adopted a slice-wise approach to address memory constraints and limited training data, as CBCT reconstructions typically involve large volumetric datasets.

\subsection{Datasets}
We used the publicly available \emph{MMDental} dataset~\cite{wang2025mmdental}, comprising 403 dental CBCT scans. Of these, 40 scans were used for validation and 40 for testing. We resized the scans to the size of $256 \times 256 \times 256$. The open-source toolbox \mbox{TIGRE}~\cite{biguri2016tigre} was used to generate $300$ projection views of size $256 \times 256$ in the range of $0^{\circ} \sim 360^{\circ}$ with an isotropic pixel size of \SI{3.0}{\milli\metre}. We set the source-to-detector distance to \SI{1500}{\milli\metre} and the detector-to-object distance to \SI{700}{\milli\metre}. Poisson and Gaussian noise were added to the projections to simulate realistic acquisition conditions. The projections were then cropped along the width dimension to create truncated projections of varying sizes, with crop widths ranging from 150 to 200\,pixels. To mimic reconstructions acquired at different small-FOV locations, the image volume was shifted by multiple physical offsets relative to the center of the original scans.
In addition, we validated the proposed method on a smaller dataset comprising 16 large FOV CBCT scans of pig jaws. More details on the CBCT units and the acquisition parameters can be found in~\cite{oliveira2025development_nota,schaub20253d}. We used one mandible for validation and one for testing. Each mandible was scanned using four different CBCT units. We resized the scans to the size of $256 \times 256 \times 256$ and generated $500$ projection views of size $512 \times 512$ in the range of $0^{\circ} \sim 360^{\circ}$ with an isotropic pixel size of \SI{1.0}{\milli\metre}. We used the same source-to-detector distance and detector-to-object distance as for the MMDental dataset. To crop the FOV, we removed 200\,pixels from the width while keeping the projection height unchanged.

\subsection{Implementation Details}
Our models are implemented in PyTorch. The INR is composed of a hash-grid encoding followed by a 4-layer ReLU mutilayer perceptron with 32 hidden units. The encoding uses 20 multiresolution levels, each storing 4-dimensional feature vectors, with a base grid resolution of 16 and a hash map containing $2^{21}$ entries. We trained the INR for 30 epochs on the MMDental dataset and 100 epochs on the pig jaw dataset using an initial learning rate of $1\times10^{-3}$, which was reduced to $1\times10^{-4}$ halfway through training. A batch size of 32,768 rays was used in each iteration.

The DDPM was trained for 150,000 iterations on the MMDental dataset and 200,000 iterations on the pig jaw dataset using a batch size of 14. We optimized all models using the Adam optimizer. More details on the empirically selected hyperparameters for the INR and diffusion model are provided in our code. All experiments were carried out on a single
NVIDIA A100 (40 GB) GPU.

\subsection{Comparing Methods}
We compared the proposed approach with several baseline methods, which are listed below. In addition, we conducted ablation studies by removing either the INR component or the diffusion-based refinement stage. We conducted the main evaluation on the MMDental dataset and evaluated only the best-performing methods on the pig jaw dataset.
\begin{itemize}

\item \textbf{Conventional Reconstruction Methods:} Three non-learning-based reconstruction methods: the analytical FDK algorithm, the iterative OS-SART algorithm and OS-SART with a TV regularization.

\item \textbf{CT-Palette:} A conditional diffusion model that predicts the full untruncated reconstruction by conditioning on both the truncated image and the mask showing the image parts to be inpainted used to generate the truncated image from the ground truth. We injected noise only into the region to be inpainted. Our implementation is based on the method proposed by CT-Palette in~\cite{liman2024diffusion}, with modifications adapted to our experimental settings.

\item \textbf{INR-based Projection Correction (INR-PC-IR):} A fast conditional 3D wavelet diffusion model~\cite{durrer2025fastwdm3d} was trained on the INR extended projections of size $256 \times 256 \times 256$. The projections generated by the diffusion model were subsequently interpolated to yield projections of size $300 \times 256 \times 256$ to compare with the ground truth. The original projection data from the uncropped FOV were incorporated into the diffusion model–corrected projections and volumetric images were subsequently reconstructed using OS-SART.

\item \textbf{Ablation 1: INR-based Iterative Reconstruction (INR-IR):} The projections rendered by the INR were reconstructed using OS-SART, without applying the diffusion-based refinement stage of our proposed method. Before the reconstruction, we copied the original projection data corresponding to the uncropped FOV into the INR-rendered projections.

\item \textbf{Ablation 2: Iterative Reconstruction-Prior DDPM (IR-DDPM):} We used the cropped ground truth projections directly as input to the OS-SART algorithm and used the resulting volumetric images as conditions to train the fast diffusion model described in Section~\ref{sec:pipeline}.

\item \textbf{IR Prior-WDM (IR-WDM):} The iterative reconstruction using cropped projections was refined using a 3D diffusion model approach based on fastWDM~\cite{durrer2025fastwdm3d} trained in the reconstructed image domain.

\item \textbf{INR Prior-WDM (INR-IR-WDM):} The INR-based iterative reconstruction (INR-IR) was refined using a 3D diffusion model approach based on fastWDM~\cite{durrer2025fastwdm3d} trained in the reconstructed image domain.

\item \textbf{INR Prior-DDPM (INR-IR-DDPM) (proposed):} The diffusion-based refinement was applied to the INR-based iterative reconstruction (INR-IR) using a fast DDPM model in 2D described in Section~\ref{sec:pipeline}.

\end{itemize}

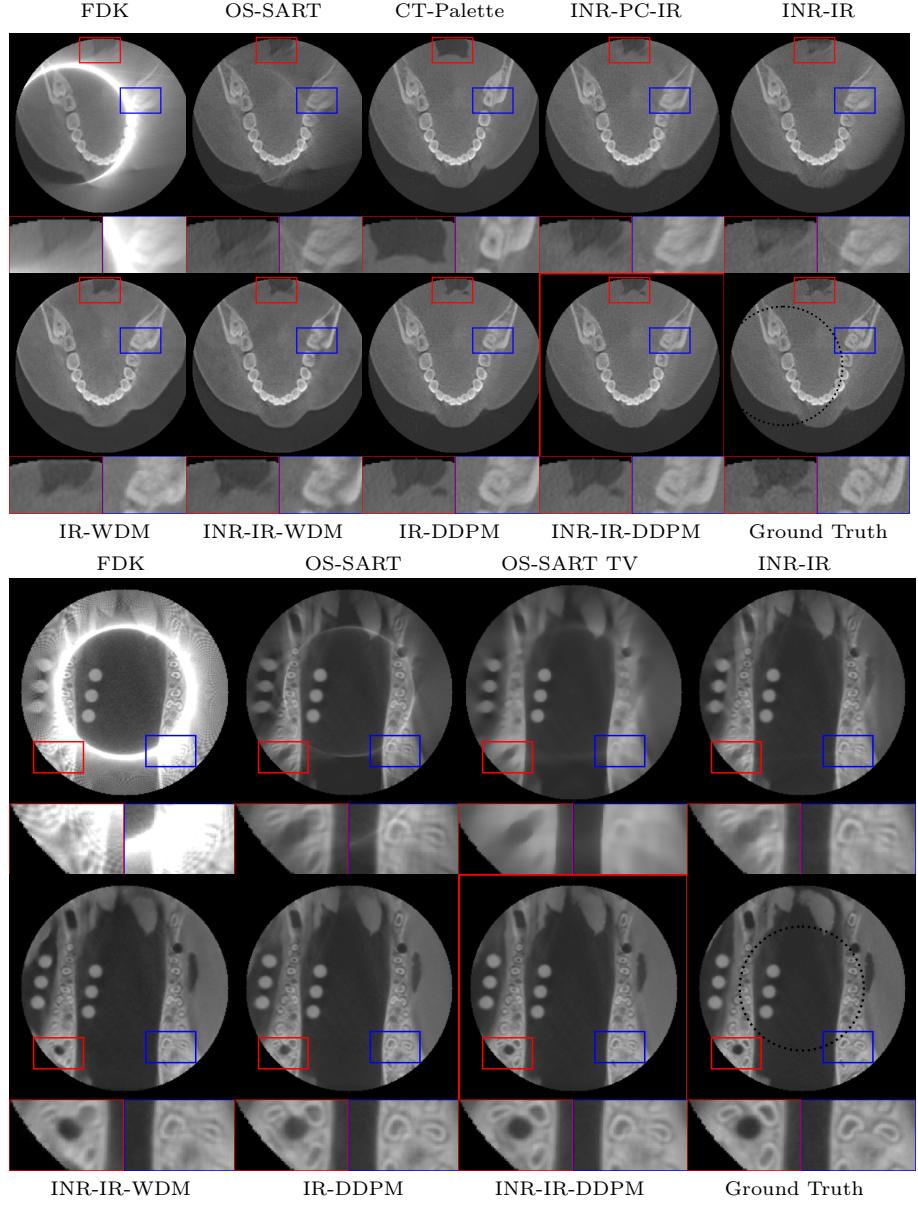
\begin{figure}
    \centering
    \scalebox{1.06}{
        \begin{tikzpicture}[scale=1.0, transform shape] 
            \node[] at (1.3, 5.4) {\scriptsize  FDK};
            \node[] at (3.4, 5.4) {\scriptsize OS-SART};
            \node[] at (5.6, 5.4) {\scriptsize CT-Palette};
            \node[] at (7.8, 5.4) {\scriptsize INR-PC-IR};
            \node[] at (10.2, 5.4) {\scriptsize INR-IR};
            \node[] at (1.3, -1.1) {\scriptsize IR-WDM};
            \node[] at (3.4, -1.1) {\scriptsize INR-IR-WDM};
            \node[] at (5.6, -1.1) {\scriptsize IR-DDPM};
            \node[] at (7.8, -1.1) {\scriptsize INR-IR-DDPM};
            \node[] at (10.2, -1.1) {\scriptsize Ground Truth};
            
            \node[] at (0, 2)   [anchor=south west]  {\includegraphics[height=3cm]{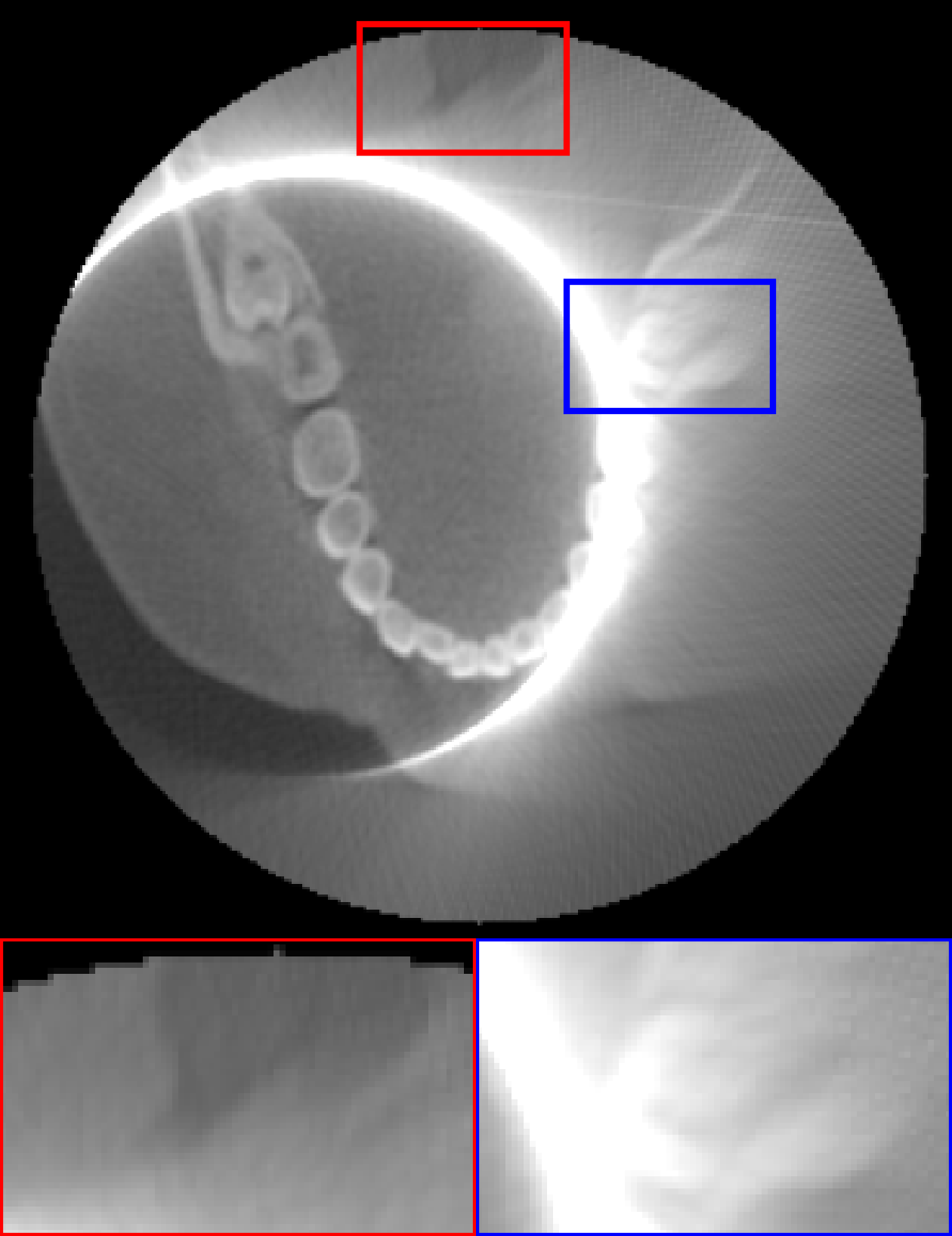}};
            \node[] at (2.2, 2)   [anchor=south west]  {\includegraphics[height=3cm]{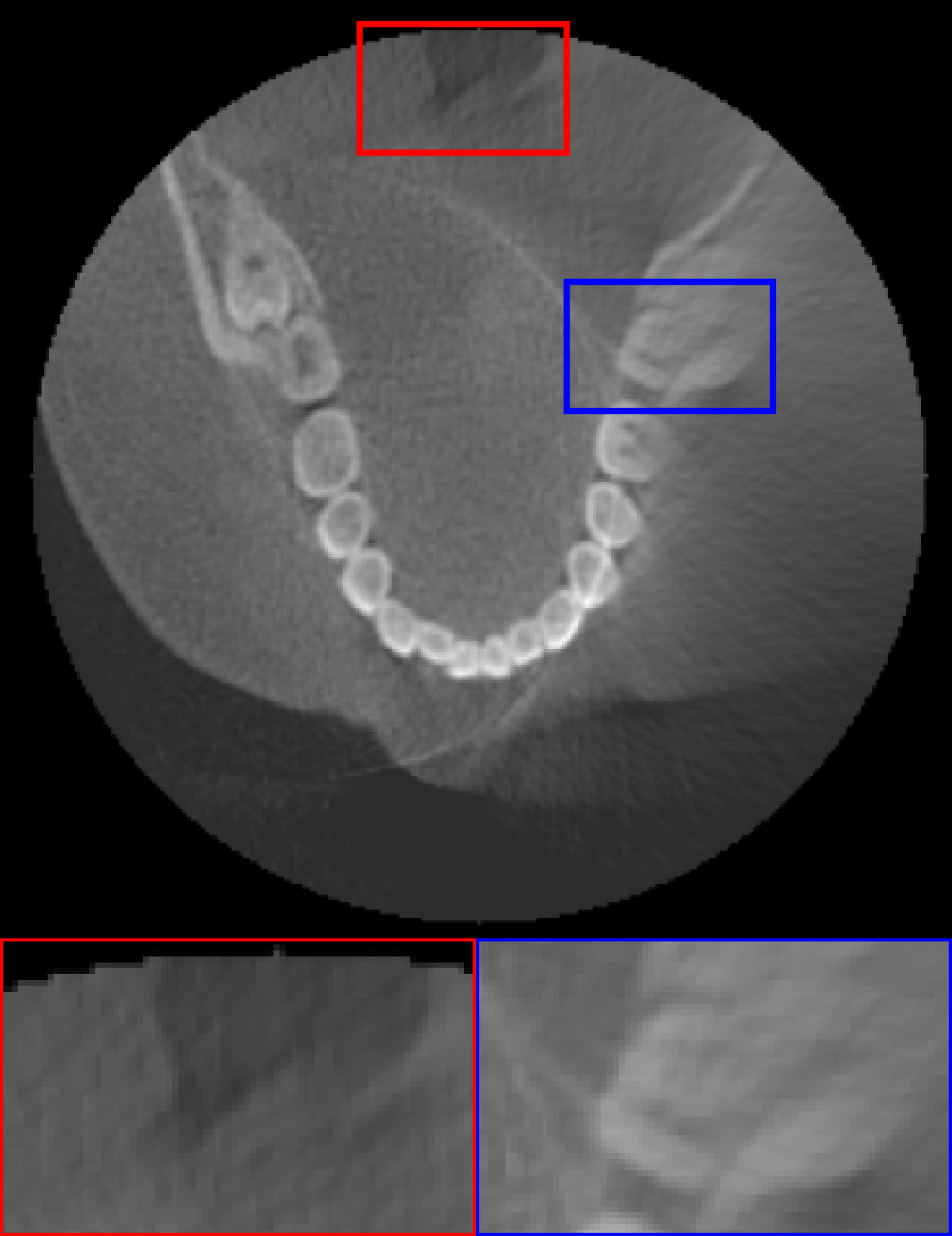}};
            \node[] at (4.4, 2)   [anchor=south west]  {\includegraphics[height=3cm]{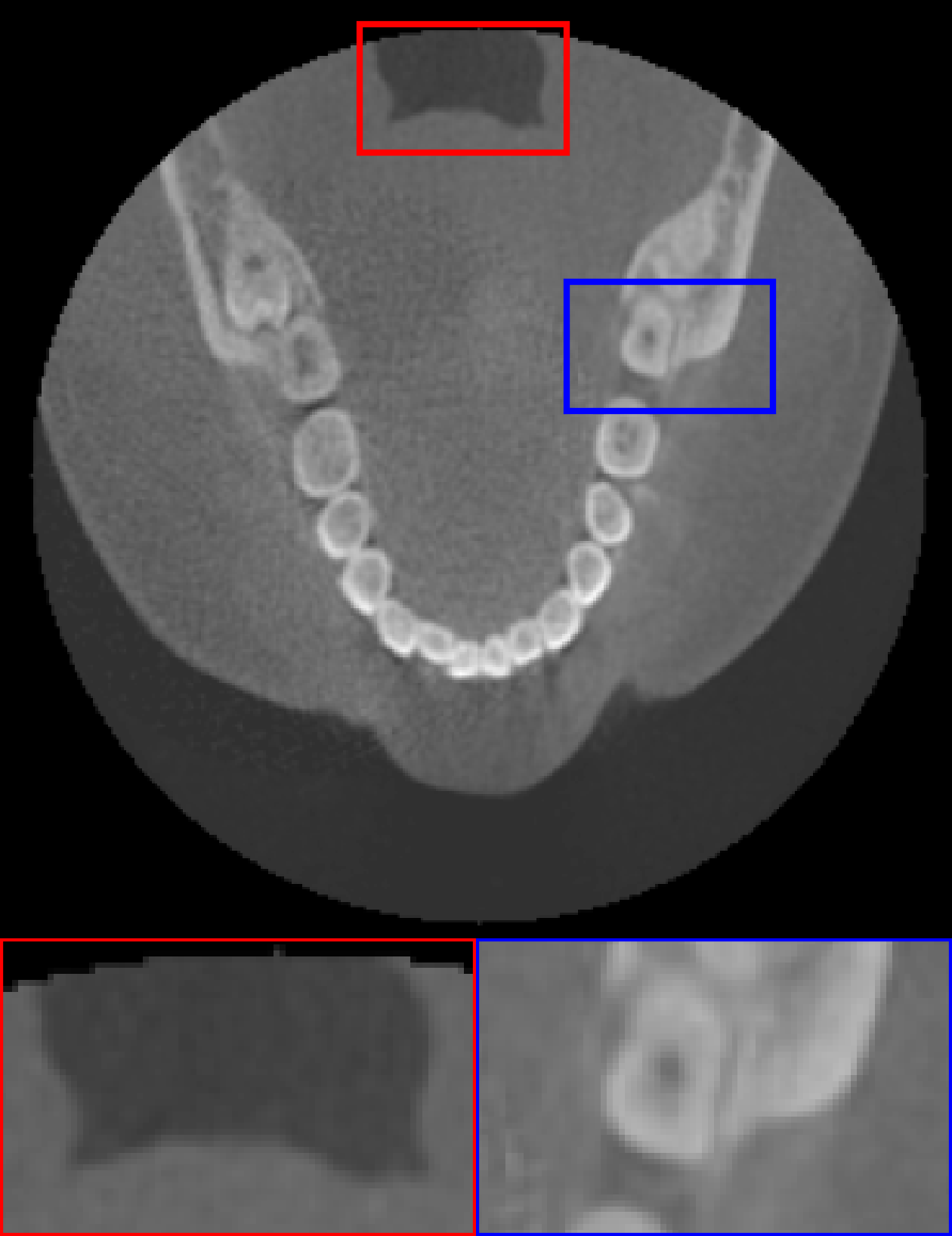}};
            \node[] at (6.6, 2)   [anchor=south west]  {\includegraphics[height=3cm]{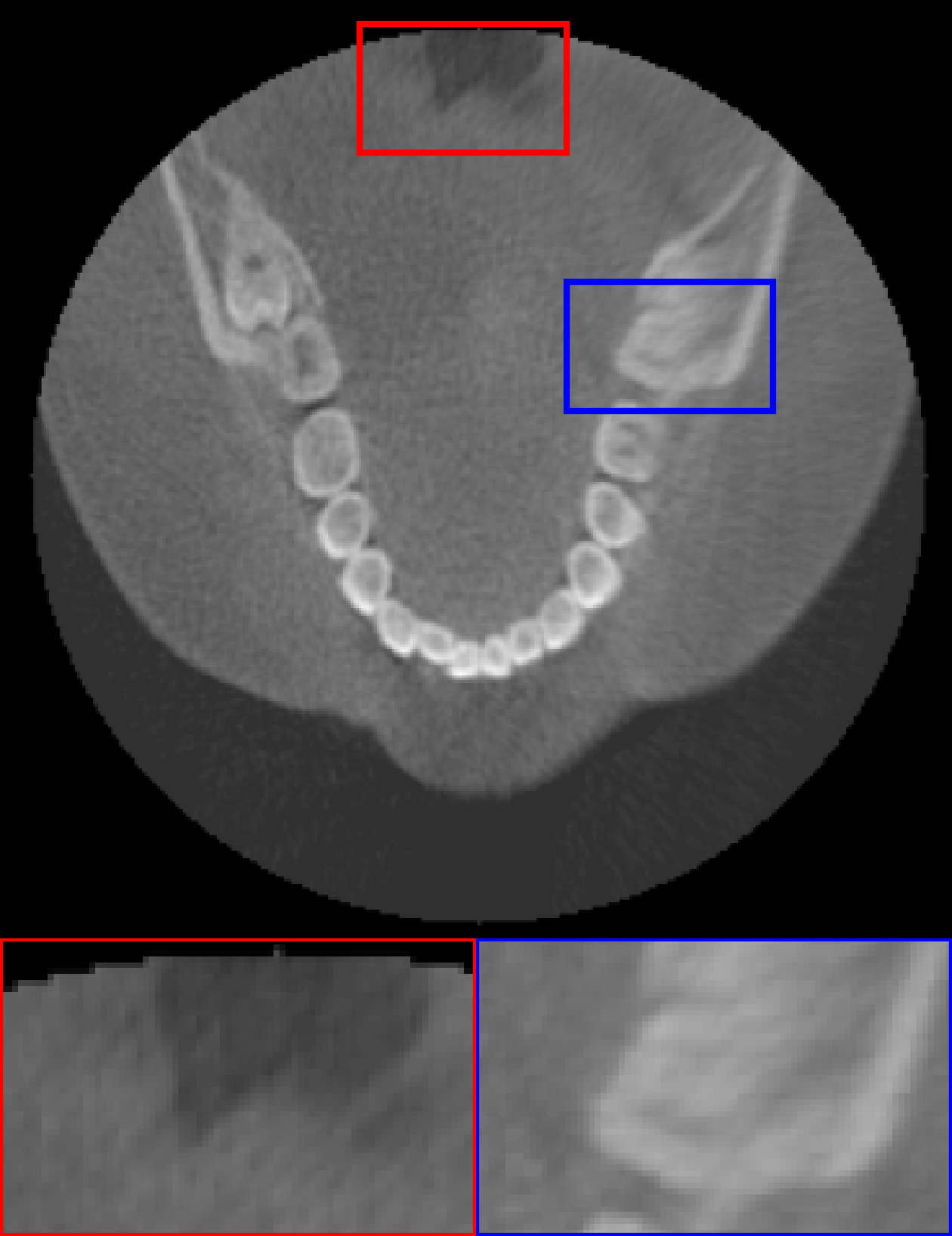}};
            \node[] at (8.91, 2)   [anchor=south west]  {\includegraphics[height=3cm]{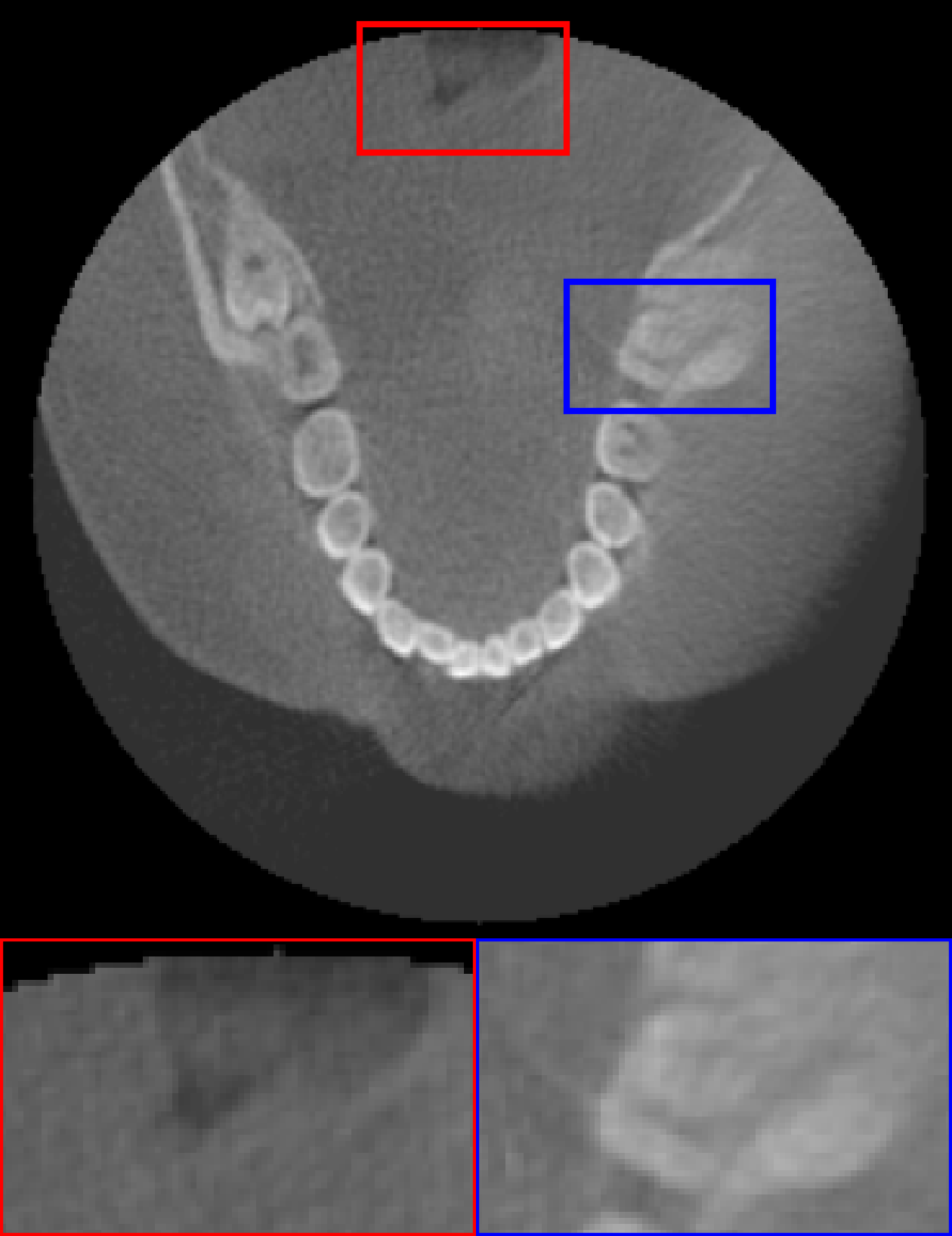}};

            \node[] at (0, -1.0)   [anchor=south west]  {\includegraphics[height=3cm]{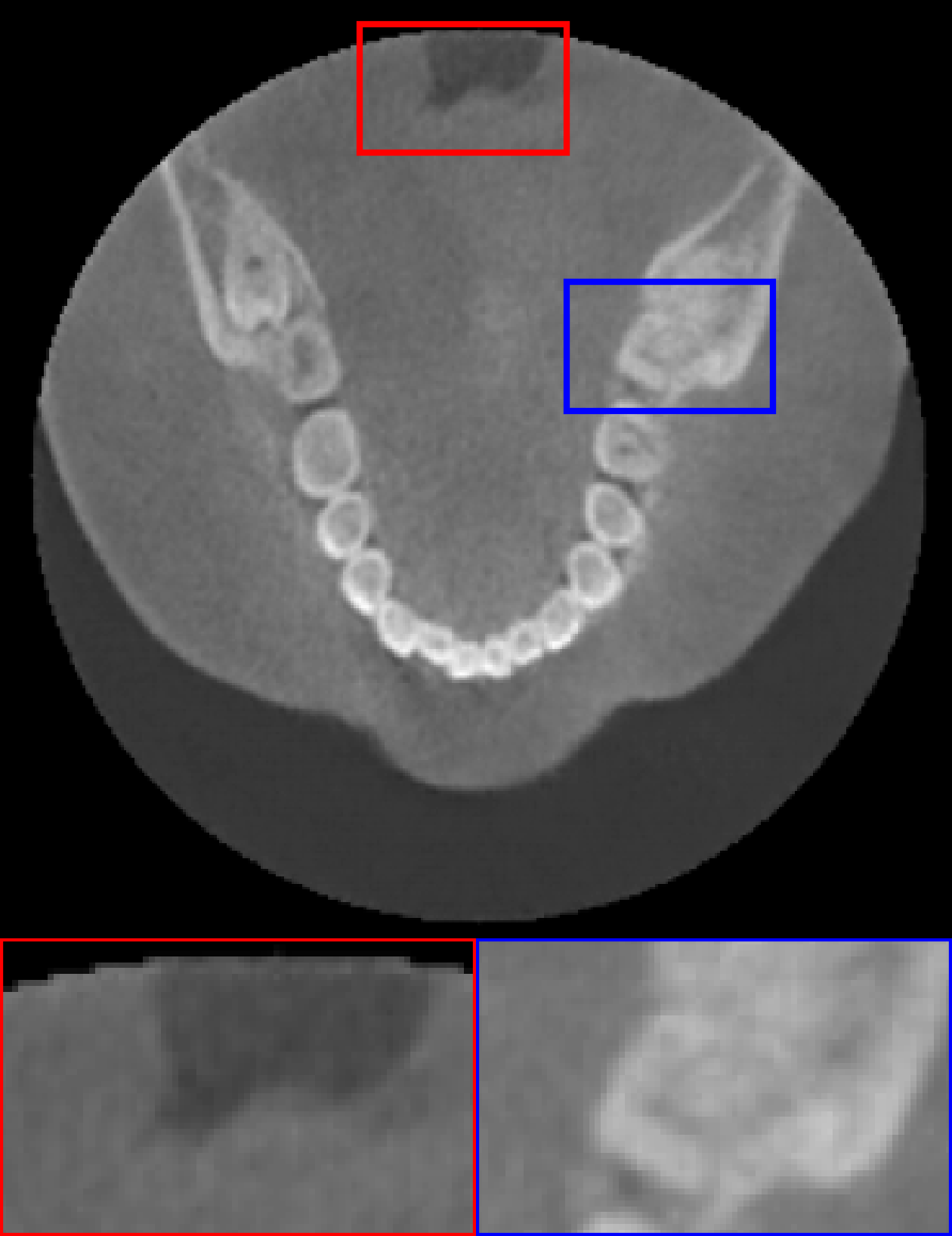}};
            \node[] at (2.2, -1.0)   [anchor=south west]  {\includegraphics[height=3cm]{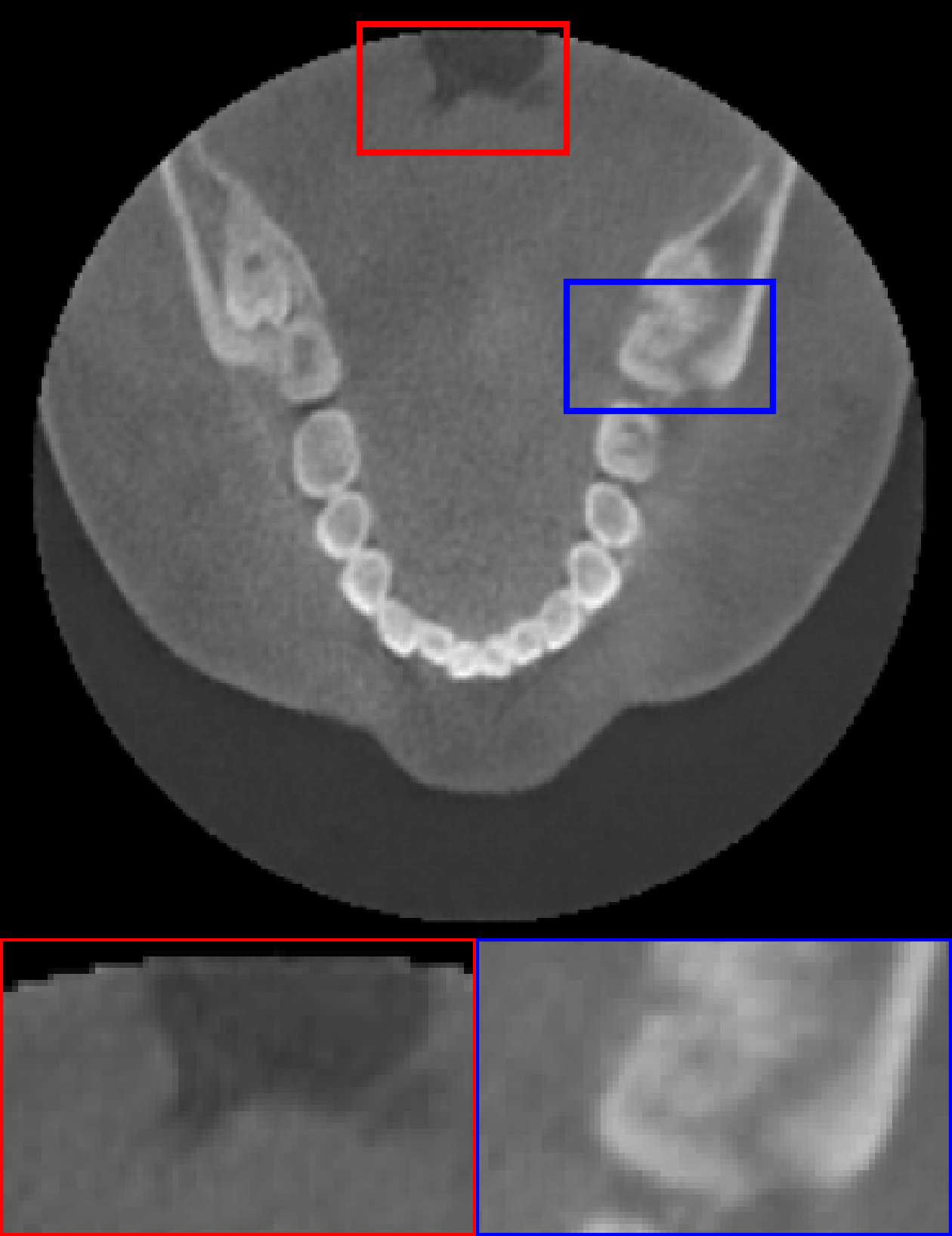}};
            \node[] at (4.4, -1.0)   [anchor=south west]  {\includegraphics[height=3cm]{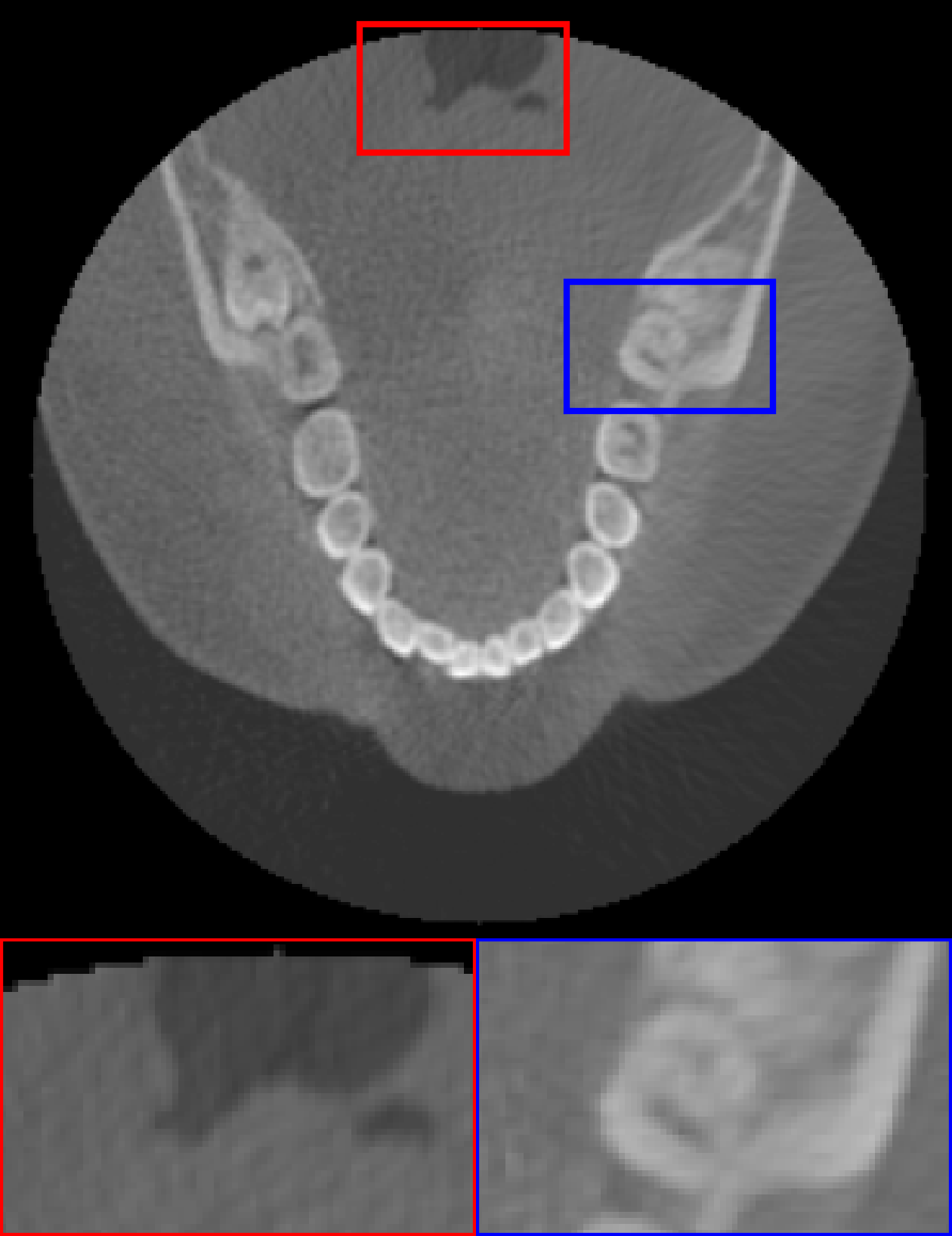}};
            \node[] at (6.6, -1.0)   [anchor=south west]  {\includegraphics[height=3cm]{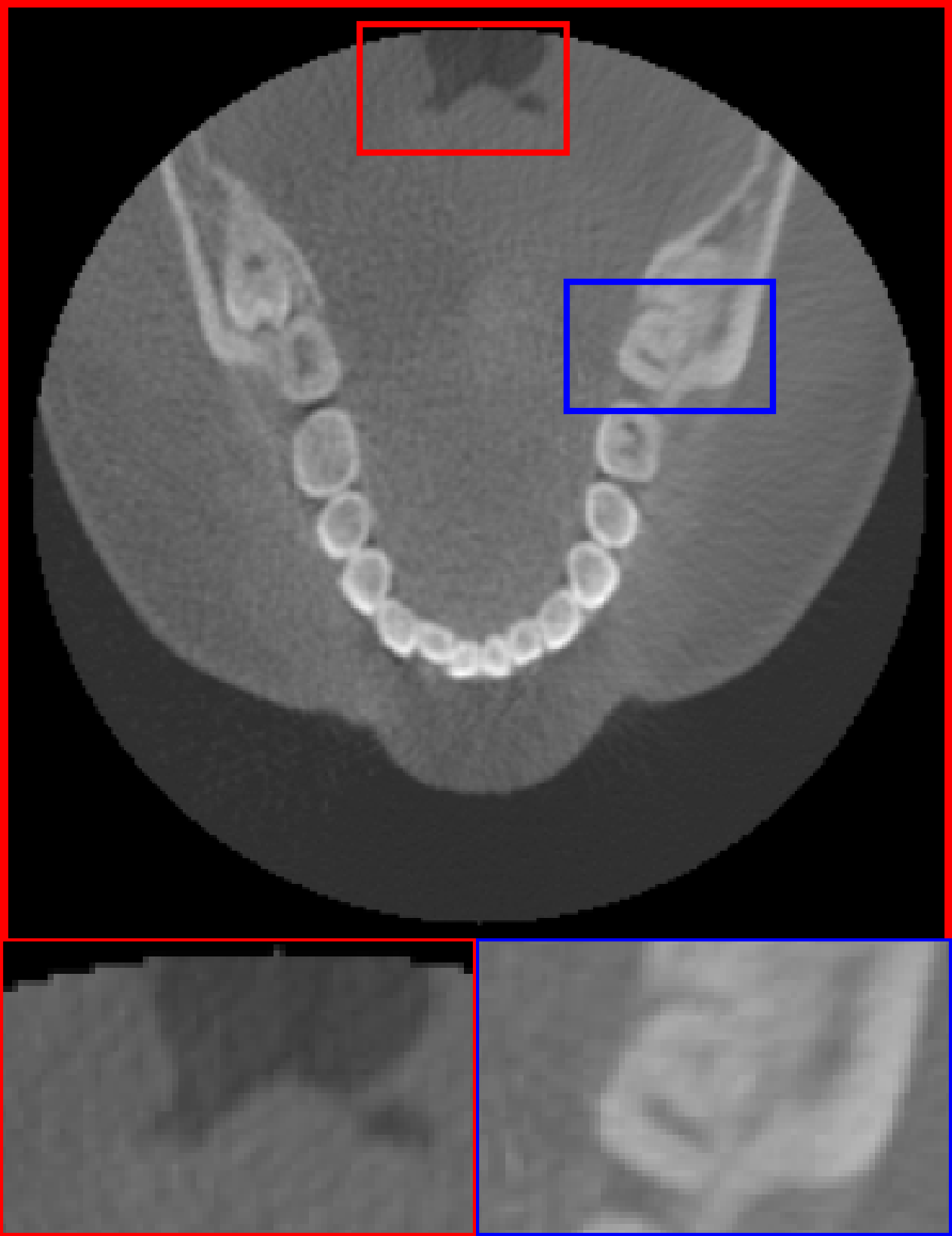}};
            \node[] at (8.91, -1.0)   [anchor=south west]  {\includegraphics[height=3cm]{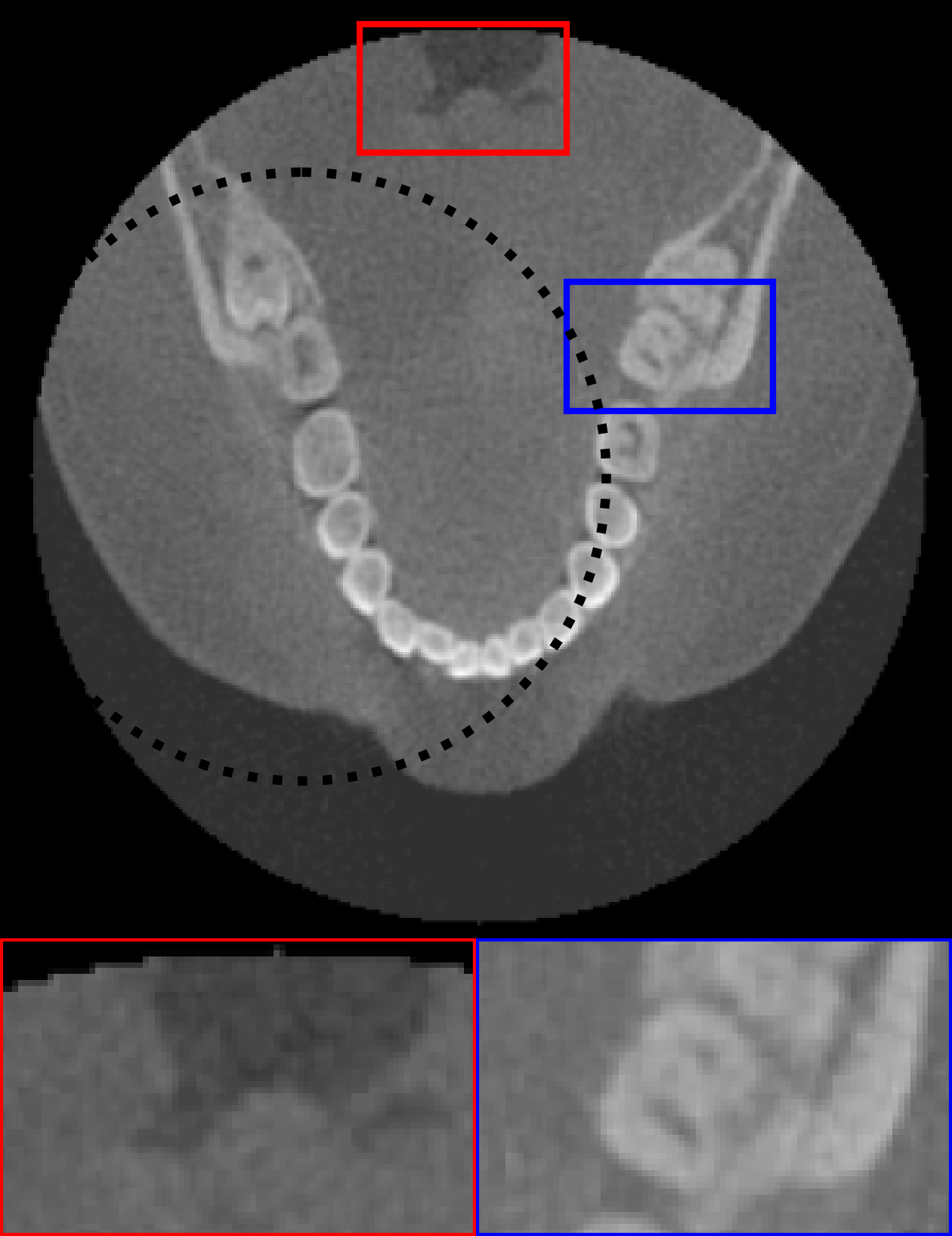}};

            \node[] at (1.5, -1.5) {\scriptsize FDK};
            \node[] at (4.4, -1.5) {\scriptsize OS-SART };
            \node[] at (7.1, -1.5) {\scriptsize OS-SART TV};
            \node[] at (9.9, -1.5) {\scriptsize INR-IR};

            \node[] at (0, -5.5)   [anchor=south west]  {\includegraphics[height=3.7cm]{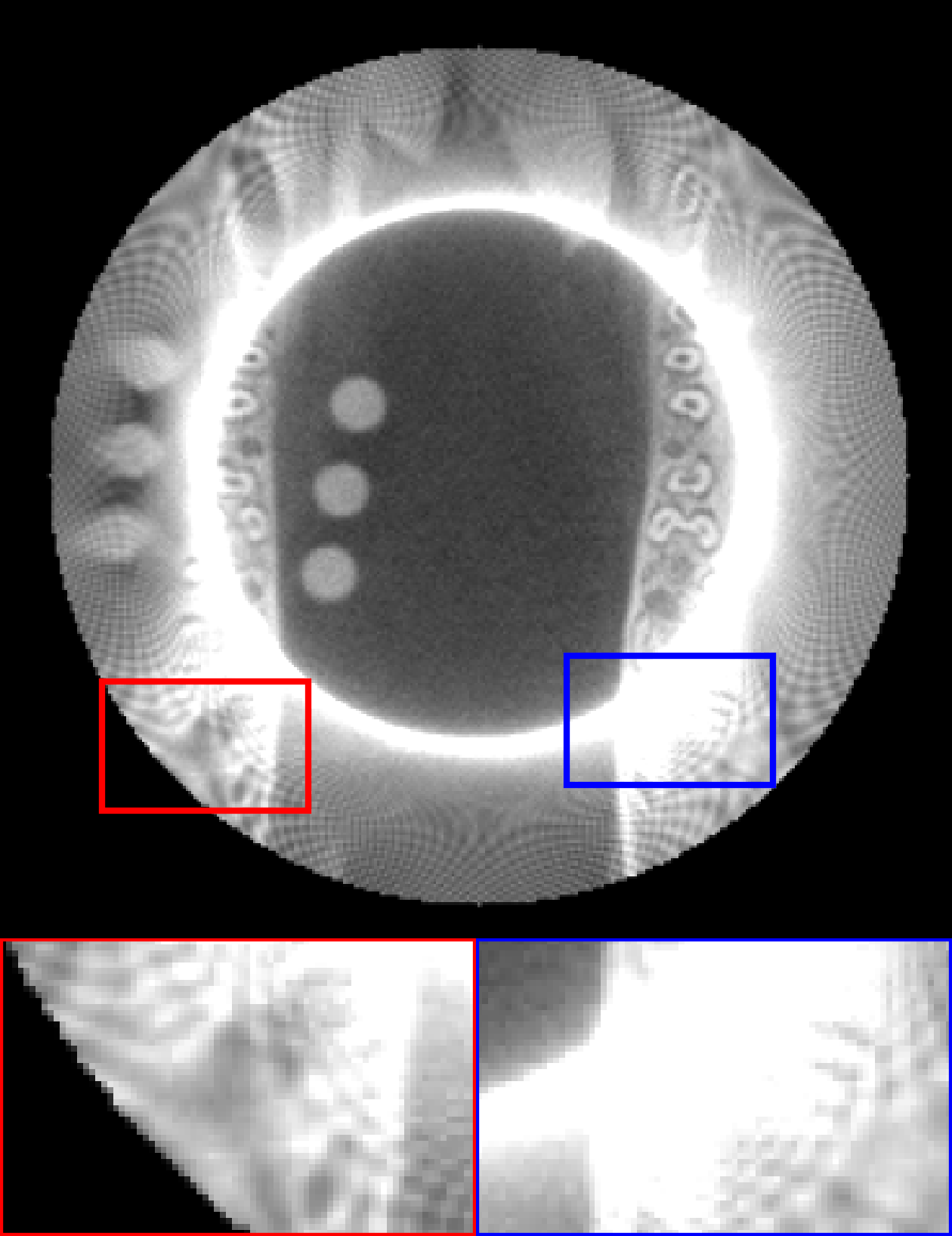}};
            \node[] at (2.8, -5.5)   [anchor=south west]  {\includegraphics[height=3.7cm]{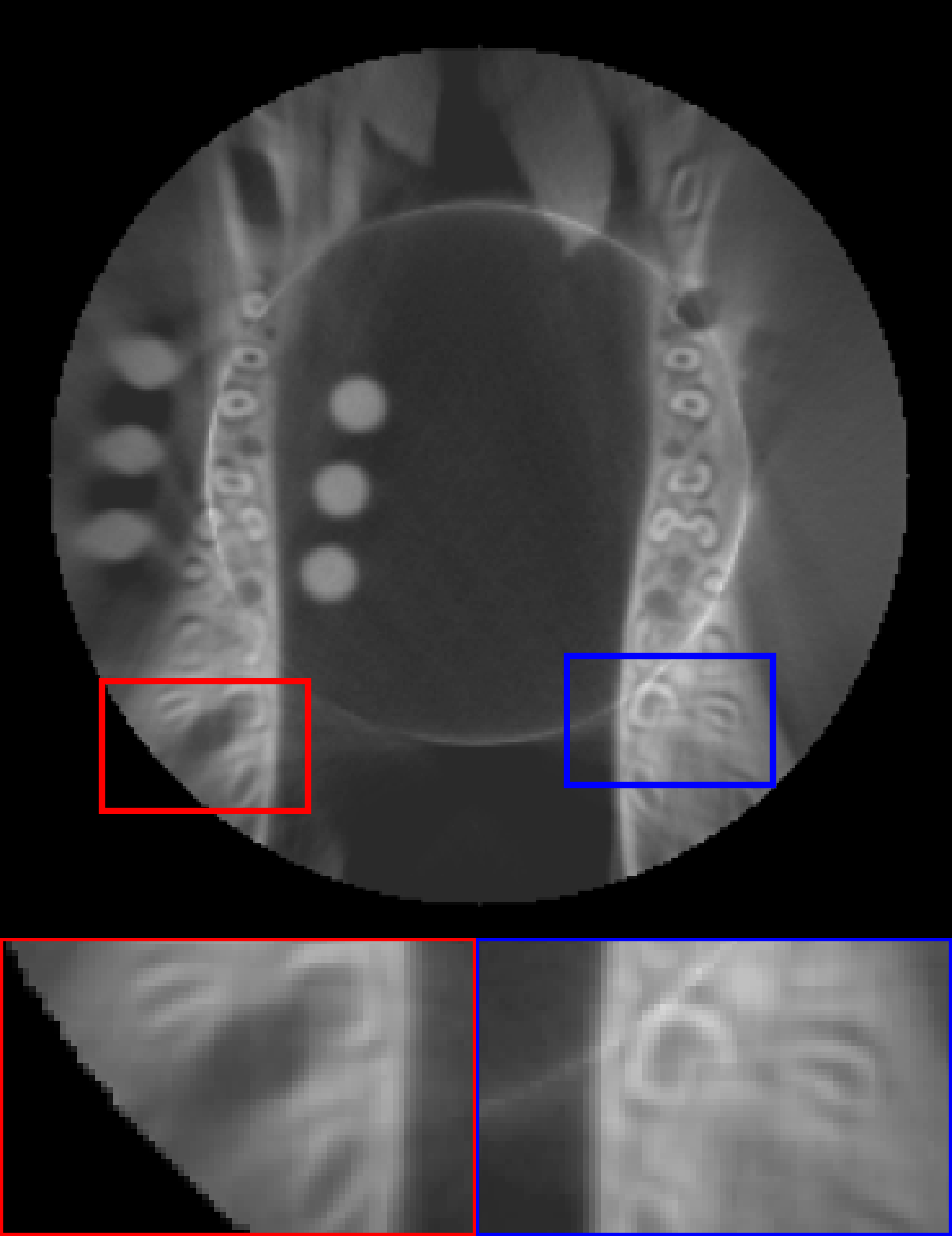}};
            \node[] at (5.6, -5.5)   [anchor=south west]  {\includegraphics[height=3.7cm]{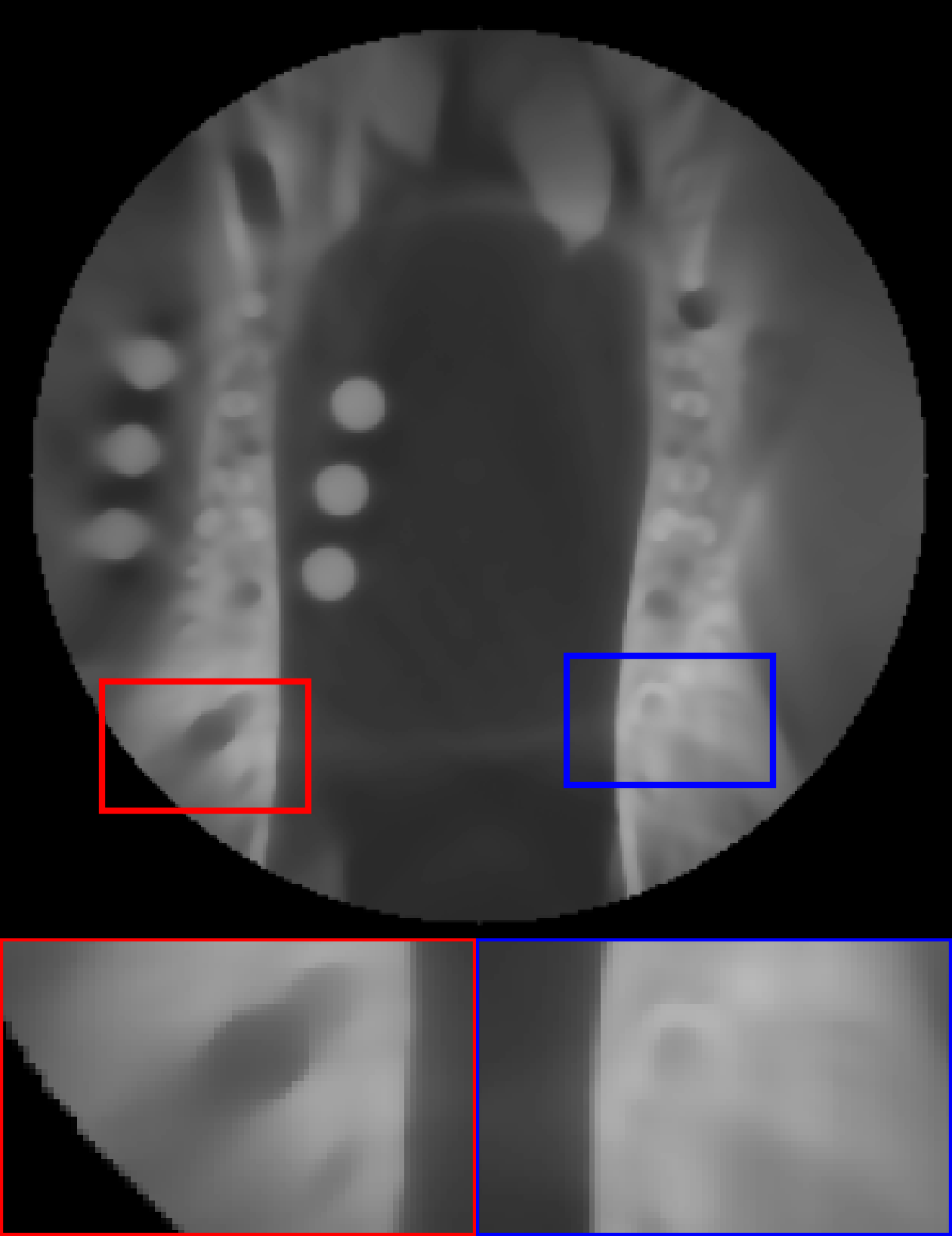}};
            \node[] at (8.45, -5.5)   [anchor=south west]  {\includegraphics[height=3.7cm]{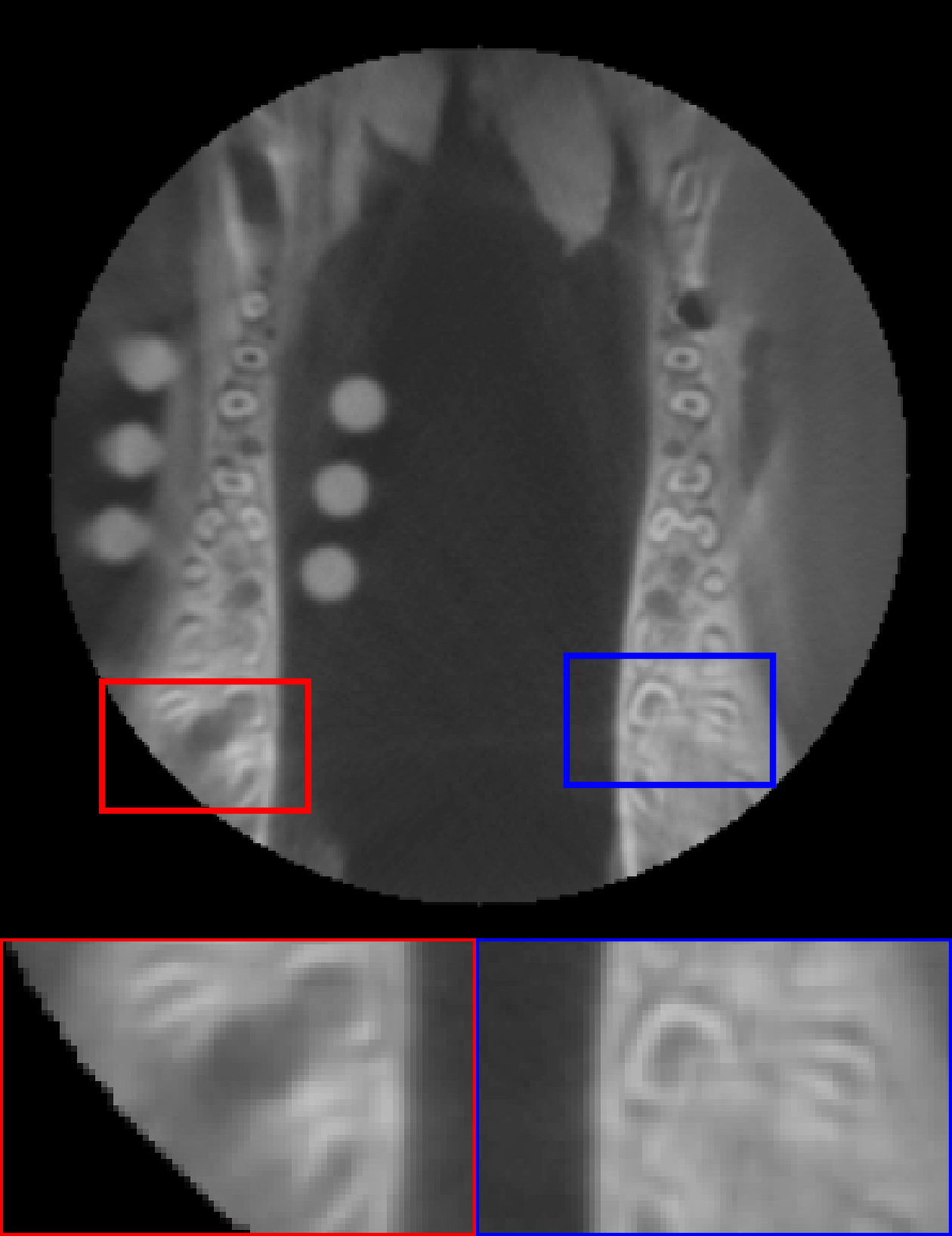}};

            \node[] at (0, -9.2)   [anchor=south west]  {\includegraphics[height=3.7cm]{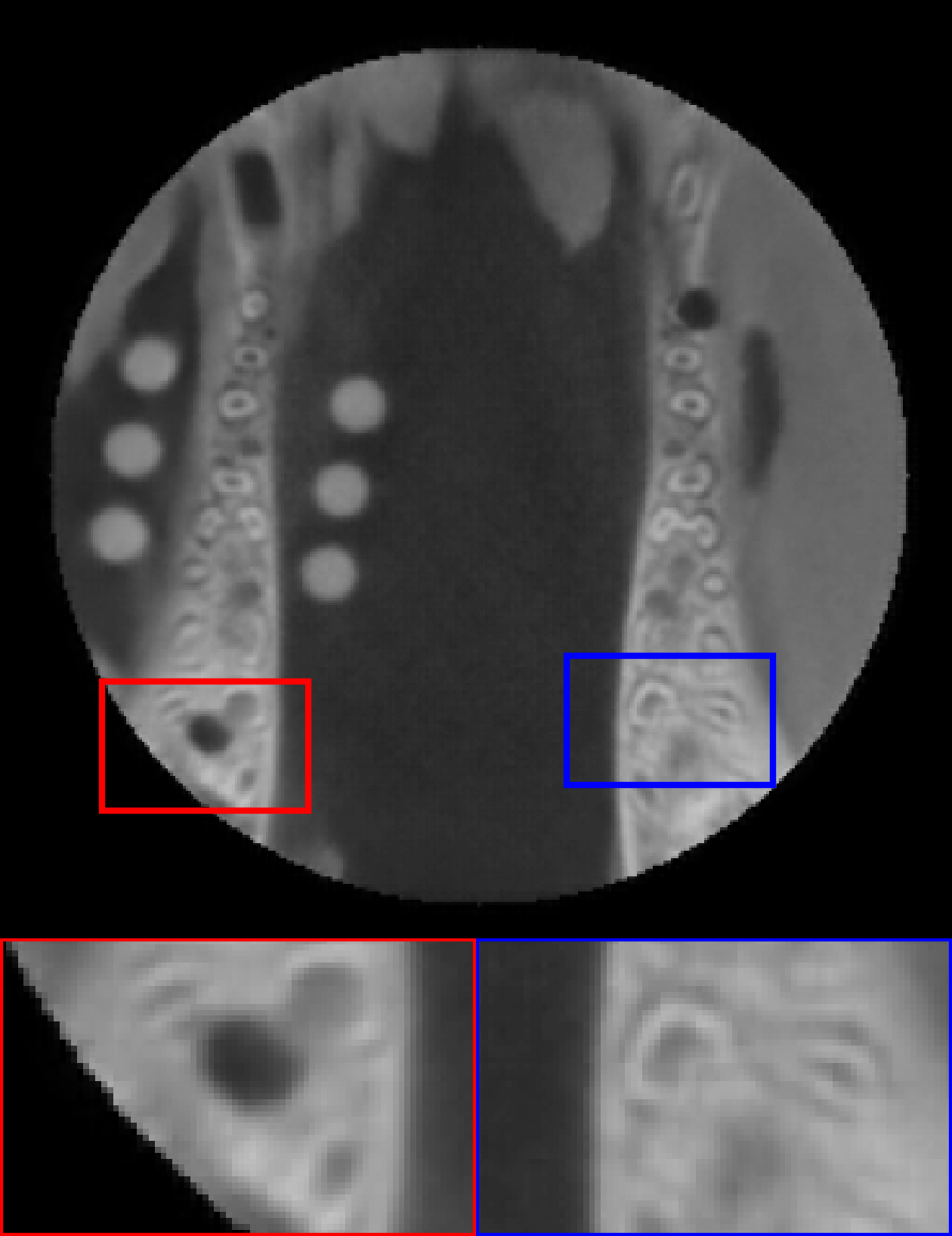}};
            \node[] at (2.8, -9.2)   [anchor=south west]  {\includegraphics[height=3.7cm]{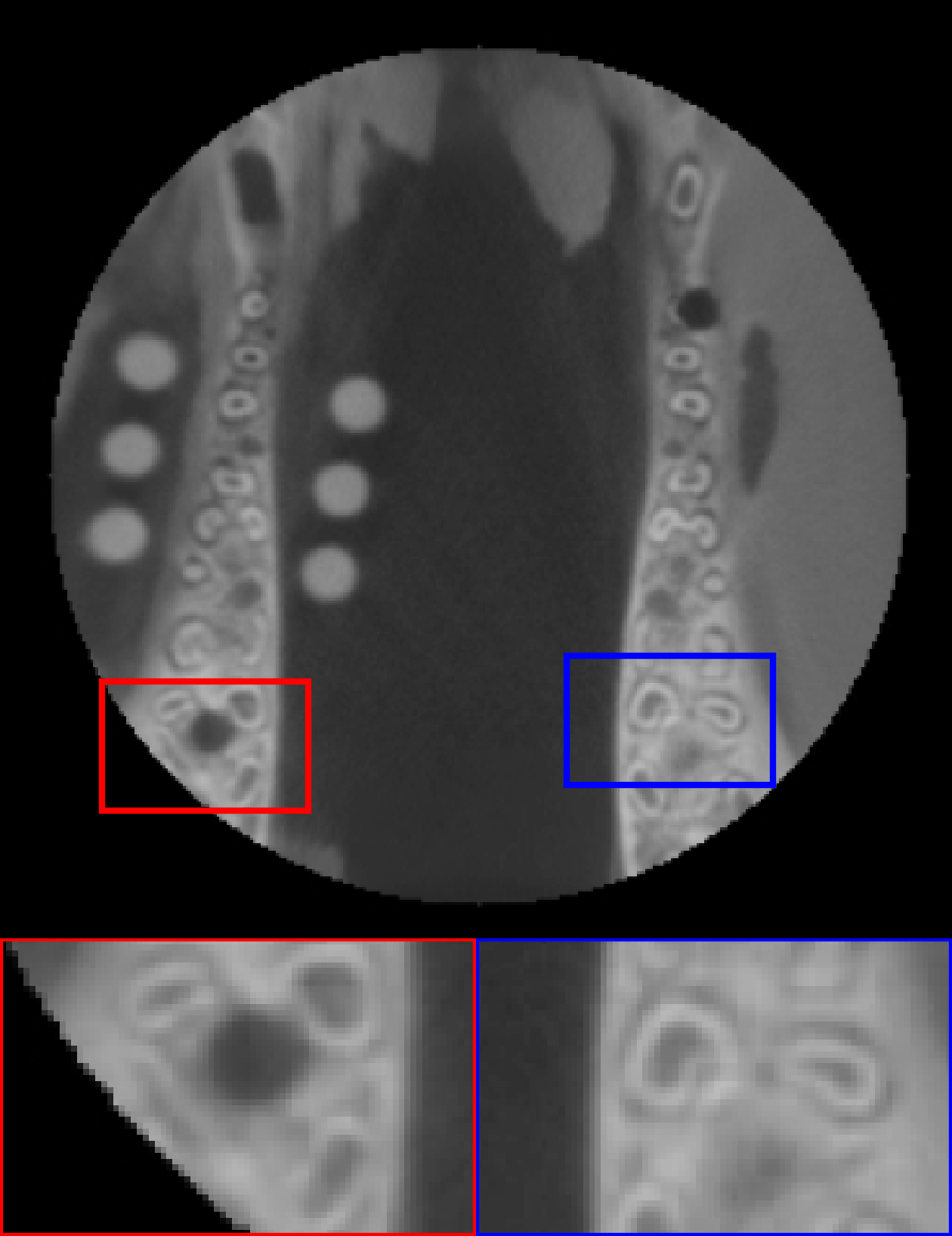}};
            \node[] at (5.6, -9.2)   [anchor=south west]  {\includegraphics[height=3.7cm]{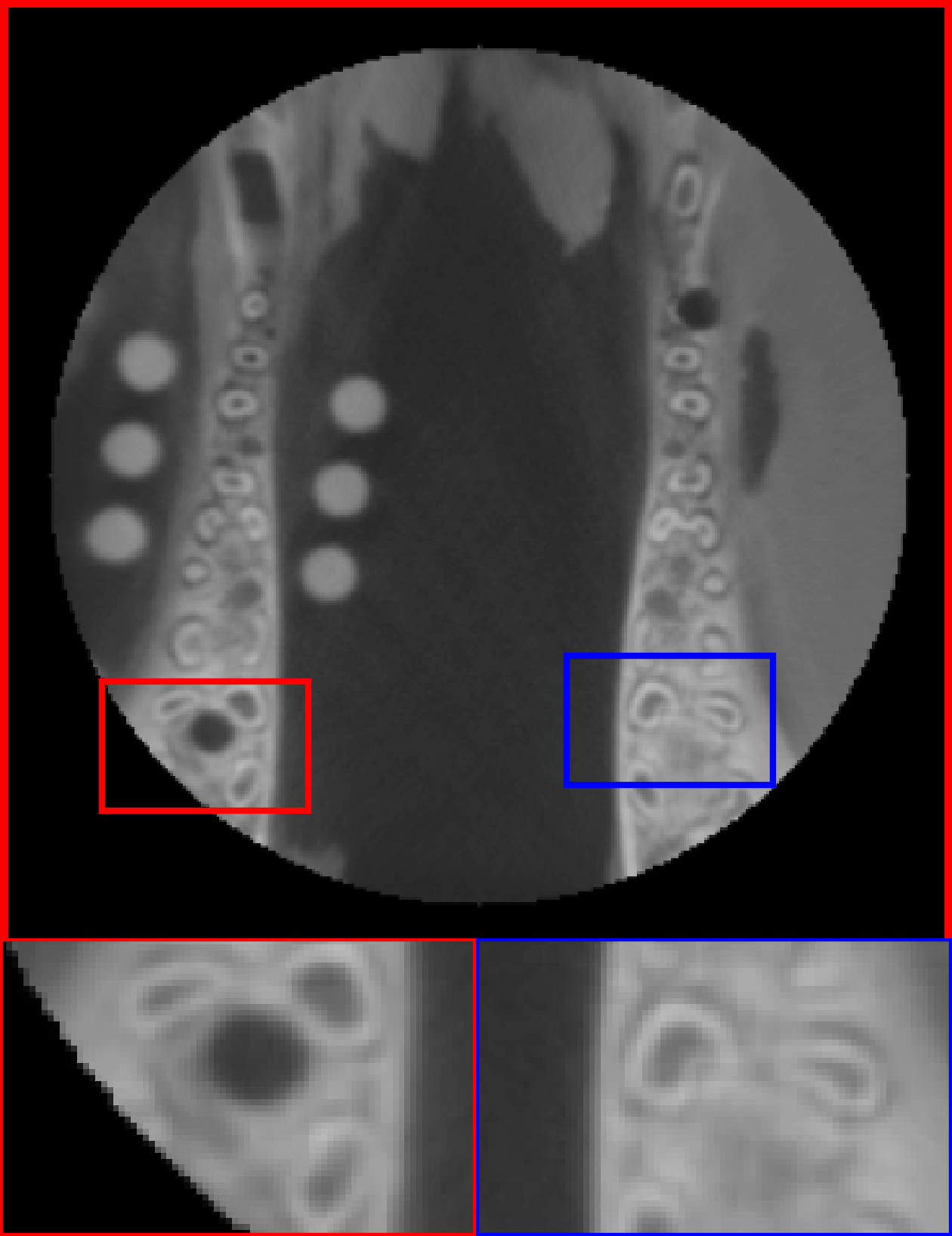}};
            \node[] at (8.45, -9.2)   [anchor=south west]  {\includegraphics[height=3.7cm]{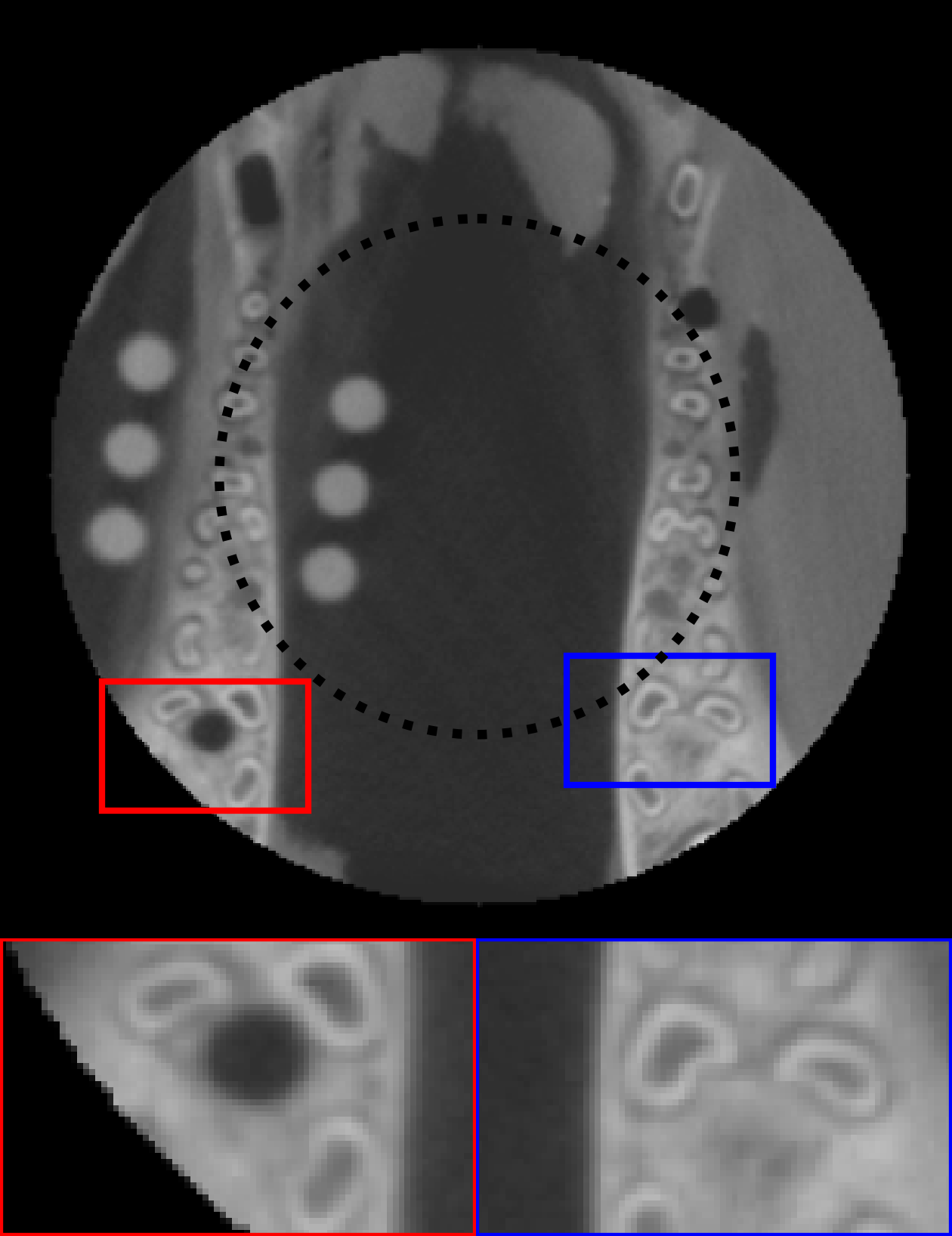}};

            \node[] at (1.5, -9.3) {\scriptsize INR-IR-WDM};
            \node[] at (4.4, -9.3) {\scriptsize IR-DDPM};
            \node[] at (7.1, -9.3) {\scriptsize INR-IR-DDPM };
            \node[] at (9.9, -9.3) {\scriptsize Ground Truth};

        \end{tikzpicture}
        }
    \caption{Qualitative comparison of the evaluated methods on the MMDental dataset (rows 1–2) and the pig jaw dataset (row 3-4). The boundary between the non-truncated and truncated FOV is indicated by a black dotted circle in the ground truth image. We show only a selection of the evaluated methods.}
    \label{comparison_MMDental_pig}
\end{figure}
\section{Results}
We quantitatively evaluated whole-image reconstructions using mean absolute error (MAE), structural similarity index measure (SSIM), peak signal-to-noise ratio (PSNR) and learned perceptual image patch similarity. 
The results are summarized in Table~\ref{tab:evaluation}. Visualization of the results is provided in Fig~\ref{comparison_MMDental_pig}. We reconstructed the ground truth with the iterative OS-SART algorithm using the uncropped projections. Compared with FDK, iterative algorithms reduce truncation artifacts but fail to accurately recover structures beyond the original FOV. Incorporating TV regularization results in blurring because the regularization suppresses high-frequency components of the image. CT-Palette hallucinates structures in the extended regions (see insets in Fig.~\ref{comparison_MMDental_pig}) and suffers from slow sampling because of the many diffusion steps. The INR-based projection correction (INR-PC-IR), performs well at modeling structures beyond the original FOV. However, the reconstructed images exhibit slight blurring.
The INR-based iterative reconstruction (INR-IR) outperforms iterative reconstructions with cropped projections; however, it fails to recover small structures outside the original FOV. The addition of the DDPM-based correction (INR-IR-DDPM) enhances the recovery of structural detail in these regions. A similar improvement is observed when the DDPM-based refinement is applied to the OS-SART reconstruction (IR-DDPM). In this case, the diffusion model successfully eliminates truncation artifacts and preserves structural detail. Replacing the 2D DDPM with 3D fastWDM yields good performance but increases blurring in the extended FOV.

We trained the DDPM model conditioned on the INR prior using 2, 3 and 1000 diffusion steps and compared their performance (see Table~\ref{tab:evaluation}). Increasing the number of diffusion steps from 2 to 3 yields a substantial improvement in performance. The highest performance is achieved by the diffusion model with 1000 diffusion steps. However, its sampling time of $\approx\SI{2.2}{\hour}$ per volume makes it impractical for clinical use. We therefore propose the model with 3 diffusion steps, which achieves the second-best performance while requiring only \SI{25}{\second}. Furthermore, Table~\ref{tab:evaluation} reports the average runtimes for the evaluated methods. For the generative models CT-Palette, fastWDM and DDPM, the reported times correspond to the generation of an entire volumetric image. For the INR, the reported time includes both the optimization of the INR until convergence and the subsequent sampling of the extended projections. The iterative algorithms required approximately 4--6 minutes per reconstruction.

\begin{table}[h!]
\centering
\caption{Mean $\pm$ standard deviation. The best and second-best results are highlighted in bold and underlined, respectively. Timesteps are denoted by T. DDPM runtime results are reported for T=2/3/1000.}
\setlength{\tabcolsep}{3pt} 
\begin{tabular}{cc|ccccc} \toprule
    & \bftab MMDental & SSIM ($\uparrow$) & PSNR ($\uparrow$) & MAE ($\downarrow$) & LPIPS ($\downarrow$) \\ \midrule
\multirow{7}{*} 
   &FDK& 0.628$\pm$0.050 & 18.28$\pm$2.16 & 0.127$\pm$0.032 & 0.273$\pm$0.039 \\
   &OS-SART& 0.891$\pm$0.066 & 34.61$\pm$4.41 & 0.020$\pm$0.011 & 0.171$\pm$0.056 \\
   &OS-SART TV& 0.867$\pm$0.056 & 34.16$\pm$3.38 & 0.021$\pm$0.009 & 0.272$\pm$0.034 \\
   &CT-Palette& 0.884$\pm$0.035 & 33.10$\pm$2.27 & 0.015$\pm$0.005 & 0.159$\pm$0.031 \\
   &INR-PC-IR& 0.954$\pm$0.020 & 39.68$\pm$2.62 & 0.009$\pm$0.003 & 0.115$\pm$0.033 \\
   &IR-WDM T=2& 0.885$\pm$0.029 & 36.51$\pm$2.27 & 0.015$\pm$0.003 & 0.143$\pm$0.022 \\
   &INR-IR-WDM T=2& 0.870$\pm$0.023 & 37.08$\pm$1.68 & 0.015$\pm$0.002 & 0.129$\pm$0.015 \\
   &INR-IR-DDPM T=3&  \uline{0.960$\pm$0.014} & \uline{41.38$\pm$2.46} &\uline{0.008$\pm$0.002} &\uline{0.100$\pm$0.021}\\ \midrule

   & \textbf{Ablations} &  &  &  &  \\
   &INR-IR& 0.933$\pm$0.035 & 37.32$\pm$3.32 & 0.012$\pm$0.005 & 0.128$\pm$0.046 \\
   &IR-DDPM T=3& 0.955$\pm$0.016 & 40.85$\pm$2.69 & \uline{0.008$\pm$0.002}& 0.108$\pm$0.022 \\
   &INR-IR-DDPM T=2& 0.885$\pm$0.024 & 37.19$\pm$1.02 & 0.016$\pm$0.001 & 0.117$\pm$0.021 \\
   &INR-IR-DDPM T=1000&\bftab0.966$\pm$0.017&\bftab41.67$\pm$2.85&\bftab0.007$\pm$0.002&\bftab0.087$\pm$0.025\\ \midrule

\multirow{8}{*} 
   & \bftab Pig Jaws & SSIM ($\uparrow$) & PSNR ($\uparrow$) & MAE ($\downarrow$) & LPIPS ($\downarrow$) \\ \midrule
   &FDK& 0.586$\pm$0.004 & 10.54$\pm$1.03 & 0.250$\pm$0.039 & 0.380$\pm$0.033 \\
   &OS-SART& 0.874$\pm$0.011 & 28.02$\pm$0.48 & 0.037$\pm$0.003 & 0.145$\pm$0.008 \\
   &OS-SART TV& 0.783$\pm$0.011 & 24.33$\pm$0.40 & 0.049$\pm$0.004 & 0.284$\pm$0.015 \\
   &INR-IR-WDM T=2& 0.955$\pm$0.023 & 38.42$\pm$1.57 & 0.010$\pm$0.002 & 0.103$\pm$0.004 \\
   &INR-IR-DDPM T=3& \bftab 0.985$\pm$0.006 & \bftab 42.69$\pm$2.14 & \bftab 0.005$\pm$0.001 & \bftab 0.040$\pm$0.004 \\ \midrule 

   & \textbf{Ablations} &  &  &  &  \\
   &INR-IR& 0.944$\pm$0.017 & 33.14$\pm$2.74 & 0.021$\pm$0.007 & 0.078$\pm$0.010 \\
   &IR-DDPM T=3& \uline{0.979$\pm$0.010} & \uline{41.06$\pm$2.11} & \uline{0.007$\pm$0.001} & \uline{0.046$\pm$0.005} \\ \midrule

   \multirow{9}{*}
  & &CT-Palette&WDM&DDPM&INR \\ \midrule
   &\bftab Average Runtime& \SI{2.2}{\hour} & \SI{3}{\second} & \SI{15}{\second}/\SI{25}{\second}/\SI{2.2}{\hour} & \SI{20}{\minute} \\ 

\bottomrule
\end{tabular}
\label{tab:evaluation}
\end{table}

\section{Discussion and Conclusion}
The INR-based reconstruction (INR-IR) provides a substantially stronger prior for the DDPM input compared to the OS-SART result, as evidenced by the quantitative evaluation. Nevertheless, the INR-IR-DDPM and IR-DDPM methods achieve comparable performance on the MMDental dataset, highlighting the strong correction capability of the DDPM. In contrast, for the pig jaw dataset, incorporating the INR prior leads to a more pronounced performance gain, indicating that the INR prior is particularly beneficial in this setting. 

Although diffusion-based corrections are fast ($\approx \SI{25}{\second}$), the prolonged INR optimization ($\approx\SI{20}{\minute}$) remains a limitation. Future work could accelerate training, e.g., through meta-learning-based initialization.

Despite strong image quality, further improvements may be achieved through end-to-end integration of neural rendering and diffusion-based correction, followed by validation on clinical scans with real projection data.

To summarize, we propose a novel three-stage framework for truncated CBCT FOV extension that consists of INR-based truncated projection extension, iterative reconstruction and diffusion model-based image correction, which formulates the task as an image-to-image translation problem. Our proposed method demonstrated strong performance, as evidenced by both the quantitative image metrics and the qualitative visual results.

\begin{credits}
\subsubsection{\discintname}
The authors have no competing interests to declare that are
relevant to the content of this article.
\end{credits}
%
%
%
%

\bibliographystyle{splncs04}
\bibliography{micad2026}
\end{document}